\documentclass[letterpaper, 10 pt, conference]{ieeeconf}  

\usepackage[T1]{fontenc}
\usepackage[utf8]{inputenc}
\usepackage{authblk}
\usepackage{graphicx}
\usepackage[spaces,hyphens]{url}
\usepackage{algorithm}
\usepackage{algpseudocode}
\usepackage{algorithmicx}
\usepackage{array}
\usepackage{mathtools}
\usepackage{subcaption}
\usepackage{amssymb}
\usepackage{amsmath}
\usepackage{tikz}
\usepackage{diagbox}
\usepackage{flushend}
\usepackage{upgreek}
\usepackage{bm}
\usepackage{array}
\DeclareMathAlphabet{\mathcal}{OMS}{cmsy}{m}{n}
\usepackage{makecell}
\usepackage{stackengine}
\usepackage{multirow}
\usepackage[symbol]{footmisc}
\usepackage{sidecap}
\usepackage{tablefootnote}
\usepackage[flushleft]{threeparttable}
\usepackage{booktabs,caption}
\usepackage{xcolor, soul}
\usepackage{hyperref}
\sethlcolor{yellow}

\DeclareMathOperator*{\argmin}{argmin}

\usepackage{booktabs}
\usepackage{multirow}
\usepackage[normalem]{ulem}
\useunder{\uline}{\ul}{}

\algnewcommand{\LeftComment}[1]{\Statex \(\triangleright\) #1}

\IEEEoverridecommandlockouts                              

\title{\LARGE \bf
Uncertainty-aware 3D Object-Level Mapping with Deep Shape Priors
}

\author{Ziwei Liao\textsuperscript{*,1}, Jun Yang\textsuperscript{*,1}, Jingxing Qian\textsuperscript{*,1}, Angela P. Schoellig\textsuperscript{1,2}, and Steven L. Waslander\textsuperscript{1}
\thanks{*Equal contribution.}
\thanks{\textsuperscript{1}The authors are with the University of Toronto Institute for Aerospace Studies and the University of Toronto Robotics Institute, Toronto, Canada.
        {\tt\footnotesize \{ziwei.liao, jun.yang, jingxing.qian, steven.waslander\}@robotics.utias.utoronto.ca}}
\thanks{\textsuperscript{2}The author is with the Technical University of Munich and the Munich Institute of Robotics and Machine Intelligence (MIRMI).
        {\tt\footnotesize angela.schoellig@tum.de}}
}

\begin{document}

\maketitle
\thispagestyle{empty}
\pagestyle{empty}

\begin{abstract}
3D object-level mapping is a fundamental problem in robotics, which is especially challenging when object CAD models are unavailable during inference. In this work, we propose a framework that can reconstruct high-quality object-level maps for unknown objects. Our approach takes multiple RGB-D images as input and outputs dense 3D shapes and 9-DoF poses (including 3 scale parameters) for detected objects. The core idea of our approach is to leverage a learnt generative model for shape categories as a prior and to formulate a probabilistic, uncertainty-aware optimization framework for 3D reconstruction. We derive a probabilistic formulation that propagates shape and pose uncertainty through two novel loss functions. Unlike current state-of-the-art approaches, we explicitly model the uncertainty of the object shapes and poses during our optimization, resulting in a high-quality object-level mapping system. Moreover, the resulting shape and pose uncertainties, which we demonstrate can accurately reflect the true errors of our object maps, can also be useful for downstream robotics tasks such as active vision. We perform extensive evaluations on indoor and outdoor real-world datasets, achieving achieves substantial improvements over state-of-the-art methods. Our code will be available at \url{https://github.com/TRAILab/UncertainShapePose}.
\end{abstract}

\section{INTRODUCTION}
3D object-level mapping is an important problem in robotics. A challenging task is to reconstruct the shape and pose of objects in the scene observed with multiple RGB-D views. Compared to traditional approaches that employ low-level geometric primitives (e.g., points and voxels)~\cite{engel2014lsd,mur2015orb,newcombe2011kinectfusion}, object-level mapping provides a rich representation of the scene and is extremely valuable for many downstream tasks, such as navigation, planning and manipulation~\cite{salas2013slam++,sucar2020nodeslam,wang2021dsp}. Early approaches require pre-scanned CAD models for each object and then construct an object-level map by estimating an object pose for each CAD model~\cite{salas2013slam++,merrill2022symmetry,deng2021poserbpf,maninis2022vid2cad}. However, these works cannot generalize to previously unseen objects. When a CAD model is unavailable, some works segment each object in the scene and reconstruct objects using multi-view depth fusion~\cite{mccormac2018fusion++,runz2018maskfusion}. These methods can reconstruct arbitrary shapes, but the reconstruction often remains incomplete as objects tend to be only partially visible during robotic operation. 

\begin{figure}[t]
\centering
  \includegraphics[width=0.85\linewidth]{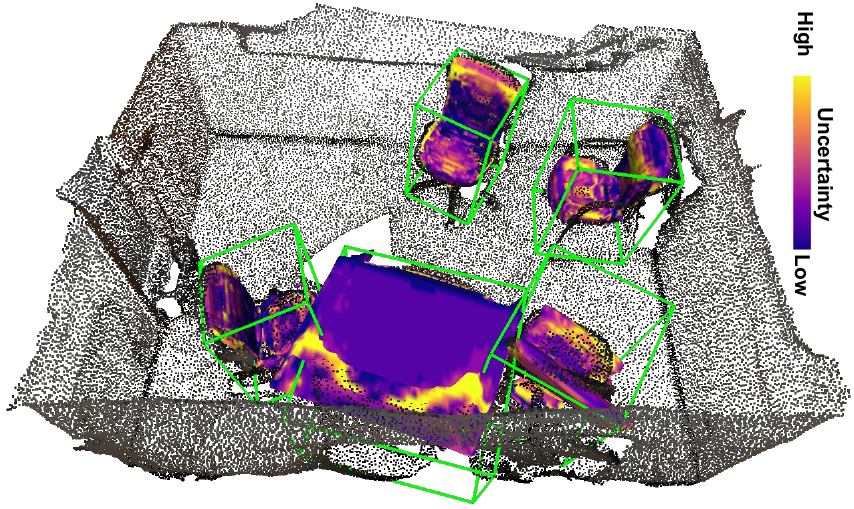}
  \vspace{-0.5\baselineskip}
\caption{Our approach takes RGB-D images as inputs and builds a 3D object-level map with dense object models, 9-DoF relative poses, and associated uncertainties.}
\vspace{-1.1\baselineskip}
\label{fig_demo}
\end{figure}

In this work, we take advantage of the recent advances in deep learning for learnt shape representation to enhance the reconstruction of the full object in object-level mapping~\cite{sucar2020nodeslam,runz2020frodo,wang2021dsp}. These approaches reconstruct objects with both dense 3D models and their relative poses. While shape priors can be used for building object-level maps for unseen objects, most methods~\cite{sucar2020nodeslam,runz2020frodo,wang2021dsp} only output a single deterministic estimation for each object's shape and pose. In contrast, for many robotic applications (e.g., robot grasping), capturing the underlying uncertainties associated with these outputs is critical. Moreover, as demonstrated in recent CAD-based works, estimating the uncertainty can improve the system performance and build high-quality object maps even in challenging scenes, where object symmetry and heavy occlusions exist~\cite{deng2021poserbpf,fu2021multi,merrill2022symmetry}.

\begin{figure*}[t]
\centering
  \includegraphics[width=0.95\linewidth]{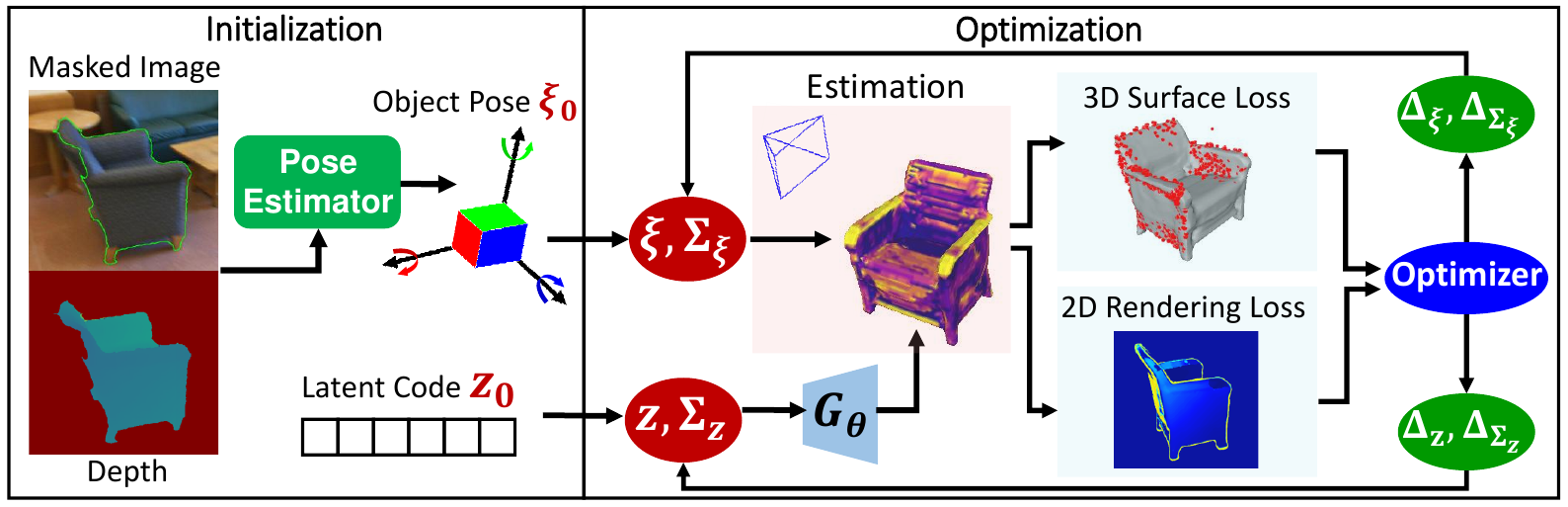}
\vspace{-0.4\baselineskip} 
\caption{An overview of the proposed uncertainty-aware object-level mapping system. We take the RGB-D images as the input and output the 3D models, 9-DoF poses, and the associated state uncertainties for the target unseen objects.}
\label{fig_pipeline}
\vspace{-0.8\baselineskip}
\end{figure*}

The above findings motivate us to build an uncertainty-aware object-level mapping system to estimate the 3D model and pose for foreground objects. For each object category, we learn its shape distributions into a latent space through a 3D generative model~\cite{park2019deepsdf}. The generative model can decode the input latent code to a detailed 3D object shape. During inference, we use the generative model to jointly optimize the latent code and the object's pose. To produce uncertainty estimates, we design a novel probabilistic optimization framework and use a combination of a 3D surface loss and a 2D rendering loss for the optimization. The final outputs of our system are dense 3D models, 9-DoF relative poses (3D translation, 3D rotation, and 3 scales along each axis), and associated state uncertainties for the target objects. We present the following key contributions  of this work:
\begin{itemize}
\item A novel probabilistic optimization framework that jointly optimizes the object's shape, pose, and associated uncertainties. Our optimization is fully differentiable and can propagate uncertainties from shape prior and object pose to different loss functions.
\item A set of two probabilistic loss functions, a 3D surface loss and a 2D rendering loss, that enable multi-view optimization of the shape prior and the relative pose, which accurately reconstructs unknown objects.
\item An uncertainty-aware object-level mapping system that can recover the 3D model, 9-DoF pose, and the state uncertainties for target unseen objects.
\end{itemize}
We demonstrate our approach's high accuracy and reliable uncertainty estimation on two public datasets. We will release the code of our approach and all baselines.

\section{RELATED WORK}
\subsection{3D Object-Level Mapping}
Prior works tackle this task by either simultaneously localizing camera poses and recovering objects, known as object-level SLAM~\cite{salas2013slam++,wang2021dsp,merrill2022symmetry,sucar2020nodeslam}, or by reconstructing the objects with known camera poses~\cite{deng2021poserbpf,maninis2022vid2cad}. Early methods build object-level maps using pre-scanned CAD models~\cite{salas2013slam++,wang2019densefusion,deng2021poserbpf, fu2021multi,maninis2022vid2cad,merrill2022symmetry}. For example, the pioneering work, SLAM++~\cite{salas2013slam++}, builds a CAD model database in advance and estimates the 6D object poses~\cite{drost2010model}. These approaches can reconstruct objects completely but cannot generalize to unknown objects outside the database. Recent works have dropped the requirement for CAD models and reconstructed the object directly by fusing multi-view depth maps~\cite{mccormac2018fusion++,runz2018maskfusion}. Although these approaches can reconstruct arbitrary objects, the reconstructed shape is usually incomplete. In our work, by exploiting a learnt shape prior, we build object-level maps with complete shapes and can generalize to unknown objects. We concentrate on the object mapping and assume camera poses are known in advance for convenience.

\subsection{3D Reconstruction with Shape Priors}
With the recent advances in deep learning, many approaches leverage learnt generative models as shape priors to reconstruct objects~\cite{sucar2020nodeslam,runz2020frodo,wang2021dsp}.
Based on neural representations at the category level~\cite{kingma2013auto,vakalopoulou2018atlasnet,mescheder2019occupancy,park2019deepsdf}, these methods can reconstruct unseen objects with complete detailed shapes. For example, NodeSLAM~\cite{sucar2020nodeslam} uses a variational autoencoder~\cite{kingma2013auto} as the shape prior and builds an object-level SLAM system capable of reconstructing full dense shapes and relative object poses. DSP-SLAM is the most closely related approach to ours. It uses DeepSDF~\cite{park2019deepsdf} as the shape prior and reconstructs the object-level map. While all these methods perform well when reconstructing unseen objects, they do not consider uncertainties underlying these estimates.

\subsection{Uncertainty Estimation in Object-Level Mapping}
In many robotic applications, it is important to estimate state uncertainties before taking action, such as robot grasping~\cite{lundell2019robust} and active perception~\cite{zhang2019beyond,yang2023active}. To this end, in many recent CAD-based approaches, the uncertainties are incorporated when estimating object poses~\cite{peng2019pvnet,okorn2020learning,deng2021poserbpf,merrill2022symmetry}. These works have demonstrated that modelling state uncertainties can significantly improve object-mapping performance. When leveraging shape priors for unseen objects, NodeSLAM~\cite{sucar2020nodeslam} develops a rendering function to incorporate the uncertainty and improve the optimization. However, NodeSLAM does not output the uncertainty for estimated object shape or pose, limiting its use cases. In our work, we explicitly estimate uncertainties for every object's shape and pose and integrate them into our optimization framework.

\section{METHODOLOGY}
\subsection{Approach Overview and Problem Formulation}
We summarize our overall framework in Figure~\ref{fig_pipeline}. Our framework takes as input the multi-view RGB-D images and builds an object-level map of a scene. For each object, we aim to estimate its 3D model (a dense mesh), $\boldsymbol{Q}_{o}$, in the object canonical coordinate frame, $O$, and a 9-DoF relative pose, $\boldsymbol{T}_{ow} \in \mathbb{R}^{4 \times 4}$, from the global (world) coordinate frame, $W$, to the object coordinate frame.

\textbf{Parametrization and Notations.} We parameterize the object's shape with an optimizable latent-shape code, $\boldsymbol{z}\in{\mathbb{R}^{64}}$, which can be passed through a decoder network, $G_\theta$, to reconstruct its 3D canonical model, $\boldsymbol{Q}_{o}$. We employ the DeepSDF~\cite{park2019deepsdf} as the shape decoder, which takes the shape code, $\boldsymbol{z}\in{\mathbb{R}^{64}}$, and a 3D query point, $\boldsymbol{p}_{o} \in{\mathbb{R}^{3}}$, under the object coordinate, $O$, as inputs, and outputs the signed distance function (SDF) value, $s$, of the 3D point:
\begin{equation}
\label{eq:deepsdf}
s = G_\theta\left(\boldsymbol{z},\boldsymbol{p}_{o}\right)
\end{equation}
The SDF represents the distance to the nearest object's surface, which can be converted to a mesh via Marching Cubes~\cite{lorensen1998marching}. The decoder network, $G_\theta$, was trained offline on a large collection of CAD models~\cite{chang2015shapenet}, and the network weights, $\theta$, are fixed during the online mapping inference. A 9-DoF pose, $\boldsymbol{T} \in \mathbb{R}^{4 \times 4}$, is constructed from a 3-DoF translation vector, $\boldsymbol{t} \in \mathbb{R}^3$, a 3-DoF rotation vector, $\boldsymbol{\phi} \in \mathfrak{so}(3)$, and a 3-DoF scaling vector, $\boldsymbol{s} \in \mathbb{R}^3$:
\begin{equation}
\label{eq:9dof}
\boldsymbol{T} = 
\begin{bmatrix}
\exp \left( \boldsymbol{\phi}^{\wedge} \right) & \boldsymbol{t}\\
\mathbf{0}^T & 1\\
\end{bmatrix}
\cdot
\begin{bmatrix}
\mathsf{diag}\left(\boldsymbol{s}\right) & \mathbf{0}\\
\mathbf{0}^T & 1\\
\end{bmatrix}
\end{equation}
where $\exp\left(\cdot\right)$ is the exponential mapping from Lie Algebra space to the corresponding Lie Group space. The operator ${\left(\cdot\right)}^{\wedge}$ converts a vector to a skew-symmetric matrix. For simplicity, we combine the translation, rotation, and scaling vectors and represent the Lie Algebra space of the 9-DoF pose, $\boldsymbol{T} \in \mathbb{R}^{4 \times 4}$, with:
\begin{equation}
\label{eq:9dof_lie_algebra}
\boldsymbol{\xi} = \left[\boldsymbol{t}, \boldsymbol{\phi}, \boldsymbol{s}\right] \in \mathbb{R}^9
\end{equation}
We assume camera poses, $\boldsymbol{T}_{wc} \in SE(3)$, are known relative to the world frame. This can be achieved using off-the-shelf SLAM methods for a hand-held camera~\cite{mur2015orb}, or robot kinematics when the camera is mounted on a robot arm~\cite{deng2020self}. 

\textbf{Uncertainty Representation.} Given object depth measurements, $\boldsymbol{D}_{1:k}$, up to viewpoint $k$, we aim to estimate the joint posterior distribution of the latent code and the object pose, $P\left(\boldsymbol{z}, \boldsymbol{\xi}_{ow}|\boldsymbol{D}_{1:k}\right)$. To represent the uncertainties, we formulate the distribution $P\left(\boldsymbol{z}, \boldsymbol{\xi}_{ow}|\boldsymbol{D}_{1:k}\right)$ with the following Gaussian distributions:
\begin{equation}
\label{eq:uncertainty_code}
    \boldsymbol{z} \sim \mathcal{N}\left(\boldsymbol{\mu}_{\boldsymbol{z}},\boldsymbol{\Sigma}_{\boldsymbol{z}}\right)
    \:\:\:,\:\:\:
    \boldsymbol{{\xi}}_{ow} \sim \mathcal{N}\left(\boldsymbol{\mu}_{\boldsymbol{\xi}_{ow}},\boldsymbol{\Sigma}_{\boldsymbol{\xi}_{ow}}\right)
\end{equation}
where $\left(\boldsymbol{\mu}_{\boldsymbol{z}},\boldsymbol{\Sigma}_{\boldsymbol{z}}\right)$ and $\left(\boldsymbol{\mu}_{\boldsymbol{\xi}_{ow}},\boldsymbol{\Sigma}_{\boldsymbol{\xi}_{ow}}\right)$ are the mean and covariance for the latent code, $\boldsymbol{z}$, and object pose, $\boldsymbol{\xi}_{ow}$, respectively. To simplify the problem, we assume each dimension of $\boldsymbol{z}$ and $\boldsymbol{\xi}_{ow}$ is independent, which allows the covariance matrices $\boldsymbol{\Sigma}_{\boldsymbol{z}}$ and $\boldsymbol{\Sigma}_{\boldsymbol{\xi}_{ow}}$ to be diagonal.

\textbf{Optimization Formulation.} We estimate all parameters, $\mathbf{X} = \bigl\{{\mu_{\boldsymbol{\xi}_{ow}},\Sigma_{\boldsymbol{\xi}_{ow}},\mu_{\boldsymbol{z}},\Sigma_{\boldsymbol{z}}} \bigl\}$, with a joint optimization formulation and solve it in an iterative manner:
\begin{equation}
\label{eq:optimization}
    \mathbf{X}^* = \argmin_{\mathbf{X}}\;\left(L_{3D} + L_{2D}\right)
\end{equation}
In our approach, we propose two probabilistic losses, a 3D surface loss, $L_{3D}$, and a 2D rendering loss, $L_{2D}$. The 3D surface loss minimizes the distance between the 3D point cloud measurement and the object surface, but is insufficient to fully constrain the object's shape and pose. As illustrated in DSP-SLAM~\cite{wang2021dsp}, the reconstructed object with 3D loss only may be much larger than its actual size in the case of partial observation. To address this issue, we introduce a novel probabilistic rendering loss function, which considers the uncertainty, to penalize shapes that grow outside the object mask and constrain its scale. Note that our optimization framework is agnostic to the particular 2D rendering function and can adapt to others~\cite{tulsiani2017multi,sucar2020nodeslam,wang2021dsp}.

In Section~\ref{sec:unc_propagate}, we first introduce a method to propagate the distribution of the latent code, $\mathcal{N}\left(\boldsymbol{\mu}_{\boldsymbol{z}},\boldsymbol{\Sigma}_{\boldsymbol{z}}\right)$, and the object pose, $\mathcal{N}\left(\boldsymbol{\mu}_{\boldsymbol{\xi}_{ow}},\boldsymbol{\Sigma}_{\boldsymbol{\xi}_{ow}}\right)$, to the SDF, $\mathcal{N}\left(\boldsymbol{\mu}_{\boldsymbol{s}},\boldsymbol{\sigma}_{\boldsymbol{s}}\right)$. Then, in Section~\ref{sec:3D_loss} and Section~\ref{sec:2D_loss}, we will describe how to use the propagated SDF distribution, $\mathcal{N}\left(\boldsymbol{\mu}_{\boldsymbol{s}},\boldsymbol{\sigma}_{\boldsymbol{s}}\right)$, to compute the 3D and 2D losses.

\subsection{Uncertainty Propagation}
\label{sec:unc_propagate}
In our mapping system, each object has its latent code distribution, $\mathcal{N}\left(\boldsymbol{\mu}_{\boldsymbol{z}},\boldsymbol{\Sigma}_{\boldsymbol{z}}\right)$, and the relative object pose distribution, $\mathcal{N}\left(\boldsymbol{\mu}_{\boldsymbol{\xi}_{ow}},\boldsymbol{\Sigma}_{\boldsymbol{\xi}_{ow}}\right)$. Given a 3D point measurement, $\mathbf{p}_{w}$, defined in the world frame, $W$, we aim to estimate its SDF distribution, $\mathcal{N}\left(\boldsymbol{\mu}_{\boldsymbol{s}},\boldsymbol{\sigma}_{\boldsymbol{s}}\right)$, in the object frame, $O$. 

For a 3D point, $\mathbf{p}_{w}$, we compute its SDF mean, $\boldsymbol{\mu}_{s}$, by transforming it to the object frame, $O$, and then passing it through the decoder network, $G_\theta$, following Equation~\ref{eq:deepsdf}:
\begin{equation}
\label{eq:sdf_mean}
\boldsymbol{\mu}_{s} = 
G_\theta\left(\boldsymbol{\mu}_{\boldsymbol{z}} \:,\: \mathbf{p}_{o}\right) =
G_\theta\left(\boldsymbol{\mu}_{\boldsymbol{z}} \:,\: \boldsymbol{\mu}_{\boldsymbol{\xi}_{ow}} \mathbf{p}_{w}\right) 
\end{equation}
To acquire the SDF variance, $\boldsymbol{\sigma}_{s}$, we linearize the entire system (Equation~\ref{eq:sdf_mean}) and propagate the covariance of the latent code, $\boldsymbol{\Sigma}_{\boldsymbol{z}}$, and object pose, $\boldsymbol{\Sigma}_{\boldsymbol{\xi}_{ow}}$. Specifically, we derive the Jacobian of the SDF value with respect to the latent code, $\boldsymbol{\mathsf{J}}_{\boldsymbol{z}}$, and object pose, $\boldsymbol{\mathsf{J}}_{\boldsymbol{\xi}_{ow}}$, as:
\begin{equation}
\label{eq:jacobian_code}
\boldsymbol{\mathsf{J}}_{\boldsymbol{z}} = \frac{\partial s}{\partial \boldsymbol{z}} = \frac{\partial G_\theta\left(\boldsymbol{z},\boldsymbol{p}_o\right)}{\partial \boldsymbol{z}}
\end{equation}
\begin{equation}
\label{eq:jacobian_pose}
\boldsymbol{\mathsf{J}}_{\boldsymbol{\xi}_{ow}} = 
\frac{\partial s}{\partial {\boldsymbol{\xi}_{ow}}} = 
\frac{\partial G_\theta\left(\boldsymbol{z},\boldsymbol{p}_o\right)}{\partial \boldsymbol{\xi}_{ow}} =
\frac{\partial G_\theta\left(\boldsymbol{z},\boldsymbol{p}_o\right)}{\partial \boldsymbol{p}_o}
\frac{\partial {\boldsymbol{p}_{o}}}{\partial \boldsymbol{\xi}_{ow}}
\end{equation}
where the terms $\frac{\partial G\left(\boldsymbol{z},\boldsymbol{p}_o\right)}{\partial \boldsymbol{z}}$ and $\frac{\partial G\left(\boldsymbol{z},\boldsymbol{p}_o\right)}{\partial \boldsymbol{p}_o}$ can be obtained via the back-propagation of the decoder network, $G_\theta$. With the linearization of Equation~\ref{eq:sdf_mean}, the SDF variance, $\boldsymbol{\sigma}_{s}^2$, is finally acquired via the following forward-propagation:
\begin{equation}
\label{eq:sdf_variance}
\boldsymbol{\sigma}_{s}^2 = 
\left[ \boldsymbol{\mathsf{J}}_{\boldsymbol{z}} \:,\:\boldsymbol{\mathsf{J}}_{\boldsymbol{\xi}_{ow}}\right]
\begin{bmatrix}
\boldsymbol{\Sigma}_{\boldsymbol{z}} & {\mathbf{0}}\\
{\mathbf{0}} & \boldsymbol{\Sigma}_{\boldsymbol{\xi}_{ow}}\\
\end{bmatrix}
{\left[ \boldsymbol{\mathsf{J}}_{\boldsymbol{z}} \:,\:\boldsymbol{\mathsf{J}}_{\boldsymbol{\xi}_{ow}}\right]}^T
\end{equation}

\subsection{Uncertainty-aware 3D Surface Loss}
\label{sec:3D_loss}
We construct the 3D surface loss, $L_{3D}$, by matching the object's depth measurements against the SDF field of the target object model. For each pixel's depth, we compute its SDF distribution, $\mathcal{N}\left(\boldsymbol{\mu}_{\boldsymbol{s}},\boldsymbol{\sigma}_{\boldsymbol{s}}\right)$, for the optimization.

Given the input depth image, $\boldsymbol{D}_k(\mathbf{u})$, from the $k^{th}$ camera frame, we first segment the object's mask, $\boldsymbol{V}_{k}$, and obtain the object's point cloud, $\mathbf{P}_{c,k}$, via back-projection:
\begin{equation}
\label{equ_back_project}
    \boldsymbol{P}_{c,k} = \Bigl\{{\boldsymbol{D}_k(\mathbf{u})} \: \boldsymbol{K}^{-1} \: \mathbf{u}^T \:,\: \mathbf{u}\in\boldsymbol{V}_k \Bigl\}
\end{equation}
where $\boldsymbol{K}$ denotes the camera intrinsic matrix and $\mathbf{u}$ represents the pixel from the object mask. The multi-view acquired point cloud is finally transformed to the global world frame, $W$, with the known camera poses, $\boldsymbol{T}_{wc,k}$:
\begin{equation}
\label{equ_pointcloud_world}
    \boldsymbol{P}_{w} = \Bigl\{ \boldsymbol{T}_{wc,k} \: \boldsymbol{P}_{c,k} \:,\: k=1:K \Bigl\}
\end{equation}
where $\mathbf{P}_{w}$ is the point cloud in the world coordinate. 

Ideally, the point cloud, $\mathbf{P}_{w}$, should perfectly align with the object surface, resulting in a zero SDF mean, $\boldsymbol{\mu}_{\boldsymbol{s}}=0$, when applying the Equation~\ref{eq:sdf_mean} on each 3D point, $\mathbf{p}_{w} \in \mathbf{P}_{w}$. For a 3D point, we measure the loss between the SDF distribution, $\mathcal{N}\left(\boldsymbol{\mu}_{\boldsymbol{s}},\boldsymbol{\sigma}_{\boldsymbol{s}}\right)$, and the target measurement (zero SDF value). Following our previous work for the shape reconstruction only~\cite{liao2023multi}, we use the Energy Score (ES), which shows to be a proper scoring rule~\cite{harakeh2021estimating}. We use it with the Monte Carlo approximation:
\begin{equation}
\label{equ:es}
\mathrm{ES}_{3D}=\frac{1}{M} \sum\limits_{m=1}^M\left\|\mathbf{s}_{m}-\overline{\mathbf{s}}\right\|  -\frac{1}{2(M-1)} \sum\limits_{m=1}^{M-1}\left\|{\mathbf{s}}_{m}-{\mathbf{s}}_{m+1}\right\|
\end{equation}
where $\overline{\mathbf{s}} = 0$ is the target SDF, and $\mathbf{s}_{m}$ denote the $m^{th}$ $i.i.d$ sample from the SDF distribution, $\mathcal{N}\left(\boldsymbol{\mu}_{s},\boldsymbol{\sigma}_{s}\right)$. We set $M=1000$ in the optimization with very little computational overhead. Considering a point cloud that has $N$ points, we compute the energy score, $\mathrm{ES}_{3D,n}$, for each 3D point measurement, and acquire its 3D loss, $L_{3D}$, with:
\begin{equation}
\label{equ:3d_loss}
L_{3D} = \frac{1}{N} \sum\limits_{n=1}^N \mathrm{ES}_{3D, n}
\end{equation}

\subsection{Probabilistic Differentiable Rendering}
\label{sec:2D_loss}
We design our 2D loss function via the differentiable SDF rendering, illustrated in Figure~\ref{fig_2d_rendering_loss}. Compared to most previous approaches~\cite{wang2021dsp,tulsiani2017multi}, which are deterministic, our rendering function, $R\left(\cdot\right)$, optimizes the object shape, pose, and state uncertainties from rendered views. It takes as input the distributions of latent code, $\mathcal{N}\left(\boldsymbol{\mu}_{\boldsymbol{z}},\boldsymbol{\Sigma}_{\boldsymbol{z}}\right)$, object pose, $\mathcal{N}\left(\boldsymbol{\mu}_{\boldsymbol{\xi}_{ow}},\boldsymbol{\Sigma}_{\boldsymbol{\xi}_{ow}}\right)$, and the known camera pose, $\boldsymbol{T}_{wc}$, and renders a depth map with uncertainties:
\begin{equation}
\label{eq:rendering_function}
    \boldsymbol{\hat{D}}_{\mu} \:,\: \boldsymbol{\hat{D}}_{\sigma} = R\left( \boldsymbol{\mu}_{\boldsymbol{z}},\boldsymbol{\Sigma}_{\boldsymbol{z}}, \boldsymbol{\mu}_{\boldsymbol{\xi}_{ow}},\boldsymbol{\Sigma}_{\boldsymbol{\xi}_{ow}},\boldsymbol{T}_{wc} \right)
\end{equation}
where $\boldsymbol{\hat{D}}_{\mu}$ and $\boldsymbol{\hat{D}}_{\sigma}$ are the rendered depth map and uncertainties from viewpoint, $\boldsymbol{T}_{wc}$. In the following sections, we describe how to obtain pixel depths and uncertainties, which will be used for computing the 2D loss.

\subsubsection{\textbf{SDF Sampling}}
\label{sec:2D_loss_1}
Following~\cite{sucar2020nodeslam,wang2021dsp}, we build our SDF renderer with differentiable ray-tracing. For each pixel,  $\mathbf{u}$, we sample $\mathcal{M}$ points uniformly along the back-projected ray, $\mathbf{r}$, within the depth range $\left[\hat{d}_{min}, \hat{d}_{max}\right]$. We denote each sampled depth with $\hat{d}_{i} = \hat{d}_{min} + \frac{i}{M}\left(\hat{d}_{max}-\hat{d}_{min}\right)$. The corresponding point under camera frame is $\mathbf{p}_i^c = \mathbf{o} + \hat{d}_{i} \mathbf{r}$, with $\mathbf{o}$ being camera optical center. For a single point, $\mathbf{p}^c$, we approximate its SDF with a Gaussian, $s \sim\mathcal{N}(\mu_s, \sigma_s^2)$, by transforming it to the world frame, $W$. To achieve better efficiency and accuracy, as in DSP-SLAM~\cite{wang2021dsp}, we only consider sampled points within a small fixed offset to the predicted surface $|\mu_s| <= \delta$.

\subsubsection{\textbf{Occupancy Probability Estimation}}
\begin{figure}[t]
\centering
  \includegraphics[width=0.87\linewidth]{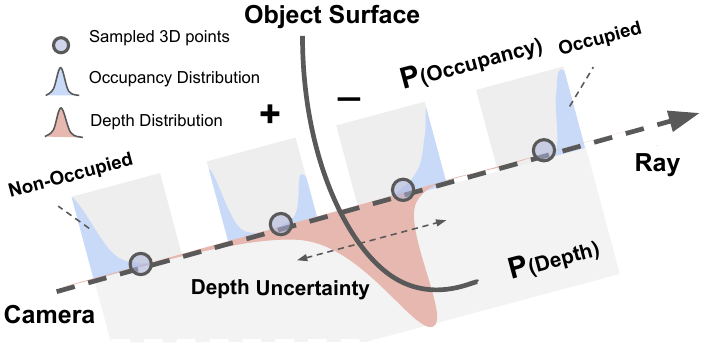}
\caption{Probabilistic differentiable rendering. We model uncertainty into the occupancy and termination probability of each sampled 3D point along a camera ray. Then, we generate a depth distribution of this ray (Sec.~\ref{sec:2D_loss}).}
\vspace{-0.5\baselineskip}
\label{fig_2d_rendering_loss}
\end{figure}

To utilize SDF in the volumetric rendering process, a common practice~\cite{wang2021dsp,sucar2020nodeslam} is to convert the SDF prediction, $s$, to an occupancy probability, $o$. In comparison, we further model the occupancy probability into a distribution. We propose to transform the SDF distribution, $s \sim\mathcal{N}(\mu_s, \sigma_s^2)$, through a mirrored sigmoid function with a slope parameter $l$:
\begin{equation}
\label{eq:sloped_sigmoid}
    o \coloneqq \textrm{sigmoid}(-ls) = \frac{1}{1+\exp(l s)}
\end{equation}
where the $l$ value encodes the cut-off threshold and controls the smoothness of the transition. We use $l=400$ in our implementation such that the cut-off threshold is around $0.01$ meters as in~\cite{wang2021dsp}. When the Gaussian SDF is passed through the sigmoid function, the resulting occupancy follows the logit-normal distribution with the following continuous density function:
\begin{equation}
\label{eq:logit_normal_pdf}
    p(o \mid \mu_s, \sigma_s^2) = \frac{1}{o (1-o)l\sqrt{2\pi\sigma_s^2}} \exp{\Bigg(-\frac{\big(\frac{-\textrm{logit}(o)}{l} - \mu_s\big)^2}{2\sigma_s^2}\Bigg)}
\end{equation}

where the logit-normal density function, $p(o \mid \mu_s, \sigma_s^2)$, is defined for $o \in (0,1)$ and $\mathtt{logit}$ represents the logit function. We illustrate the logit-normal occupancy distribution in Figure~\ref{fig_2d_rendering_loss}.

\subsubsection{\textbf{Termination and Escape Probability}}
When tracing points along the ray, $\mathbf{r}$, the ray either terminates on a surface point or escapes without hitting any point. Following previous works~\cite{sucar2020nodeslam,wang2021dsp}, we compute the termination probability, $\phi_i$, for each sampled point, $\mathbf{p}_i^c$, and the escape probability, $\phi_{M+1}$, for a camera ray, $\mathbf{r}$:
\begin{equation}
\label{eq:terminamte_escape}
\begin{aligned}
    &p(\phi_i) = p(o_i\mid \mu_{s,i}, \sigma_{s,i}) \prod^{i-1}_{j=1}{p(1-o_{j}\mid \mu_{s,j}, \sigma_{s,j})}\\
    &\quad\quad= p(o_i\mid \mu_{s,i}, \sigma_{s,i}) \prod^{i-1}_{j=1}{p(o_j\mid -\mu_{s,j}, \sigma_{s,j})}, i=1, \dots, M \\
    &p(\phi_{M+1}) = \prod^{M}_{j=1}{p(1-o_j\mid \mu_{s,j}, \sigma_{s,j})} = \prod^{M}_{j=1}{p(o_{j}\mid -\mu_{s,j}, \sigma_{s,j})}
\end{aligned}
\end{equation}
where $p(1-o_j \mid \mu_s, \sigma_s) = p(o_j \mid -\mu_s, \sigma_s)$ represents the symmetry property of logit-normal distributions. Since the logit-normal distribution is not closed under multiplication, we propose to approximate the product with a Beta distribution, $q(\phi_i)$, by matching the first and second moments. Specifically, given the quantile function, $Q(\cdot)$, of the logit-normal distribution with slope $l$:
\begin{equation}
\label{eq:logit_normal_quantile}
    Q(o_i \mid \mu_{s,i}, \sigma_{s,i}) = \textrm{sigmoid}\left(l \sqrt{2\sigma_{s,i}^2} \textrm{erf}^{-1}(2o_i-1) - l\mu_{s,i}\right),
\end{equation}
where $\textrm{erf}$ is the error function, we estimate the event posterior mean $\mu_{\phi,i}$, $\mu_{\phi,M+1}$ and variance $\sigma_{\phi,i}$, $\sigma_{\phi,M+1}$ from the SDF means and variances via Quasi-Monte Carlo with the inverse transform method~\cite{rubinstein2016simulation,raychaudhuri2008introduction}. We use 128 samples from the Sobol sequence in our implementation.

\subsubsection{\textbf{Depth Estimation and 2D Loss Term}}
\label{sec:depth_est_2D_loss}
With the estimated event probabilities, we acquire the depth distribution by computing the expectation, $\hat{\mu}_{d}$, and variance, $\hat{\sigma}_{d}^2$, over the sampled points. As in 3D loss, for each pixel $\mathbf{u}$, with the rendered depth mean $\hat{\mu}_{d}$, and variance, $\hat{\sigma}_{d}^2$, we compute its 2D loss, $\mathrm{ES}_{2D, \mathbf{u}}$, using the energy score (Equation~\ref{equ:es}). The final rendering term is defined as:
\begin{equation}
\label{equ:2d_loss}
L_{2D} = \frac{1}{|\boldsymbol{V}_s|} \sum\limits_{\mathbf{u}\in \boldsymbol{V}_s} \mathrm{ES}_{2D, \mathbf{u}}
\end{equation}
where $\boldsymbol{V}_s = \boldsymbol{V}_{o} \cup \boldsymbol{V}_{b}$ is the union of object surface pixels, $\boldsymbol{V}_{o}$, and background pixels, $\boldsymbol{V}_{b}$. Object surface pixels, $\boldsymbol{V}_{o}$, are the set of pixels from the object's mask. The background pixels, $\boldsymbol{V}_{b}$, are not on object surfaces but inside the object's 2D bounding box. Following~\cite{sucar2020nodeslam,wang2021dsp}, we assign background pixels a depth of $\hat{d}_{M+1} = 1.1\hat{d}_{max}$.

\begin{figure}[t]
\centering
  \includegraphics[width=0.92\linewidth]{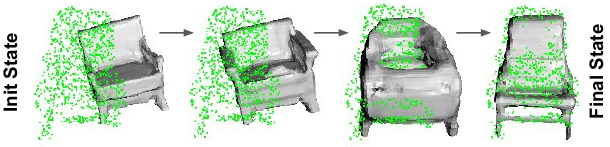}
  \vspace{-0.6\baselineskip}
\caption{The optimization process of the shape and pose of a chair instance from ScanNet with different iterations.}
\vspace{-0.8\baselineskip}
\label{fig_example}
\end{figure}

\subsection{Optimization}
Our final loss is the weighted sum of the 3D loss, $L_{3D}$, 2D losses, $L_{3D}$ and a shape code regularization term, $||\mathbf{z}||$:
\begin{equation}
\label{eq:final_loss}
    \mathbf{L}_{final} = \lambda_{s}L_{3D} + \lambda_{r}L_{2D} + \lambda_{c}{||\mathbf{z}||}^2
\end{equation}
where $\lambda_{s}$, $\lambda_{r}$ and $\lambda_{c}$ are the weights for each loss term. In our approach, we initialize the shape prior with a code $\mathbf{z} = \mathbf{0}$. The initial object pose, $\mathbf{T}_{ow}$, is obtained by matching the initial object shape (corresponds to the code $\mathbf{z} = \mathbf{0}$) to the object point cloud with the ICP algorithm. Note that the pose initialization can be replaced with any 3D object detector. We initialize the covariance matrix for shape code, $\Sigma_{\mathbf{z}}$, and object pose, $\Sigma_{\boldsymbol{\xi}_{ow}}$, by placing a constant value on their diagonal elements. In our implementation, we set $1\times e^{-6}$ and $1\times e^{-4}$ for $\Sigma_{\mathbf{z}}$ and $\Sigma_{\boldsymbol{\xi}_{ow}}$, respectively. In our work, we use the Adam optimizer~\cite{kingma2014adam} to solve the optimization problem. We use $0.005$ learning rate and run the optimizer for $200$ iterations. Figure~\ref{fig_example} shows an example of our optimization with different iterations.

\section{EXPERIMENTS}
We evaluate our framework on two real-world datasets, indoor ScanNet~\cite{dai2017scannet} and outdoor KITTI-3D~\cite{geiger2012we}. The ScanNet dataset provides RGB-D video sequences of multiple objects in complex indoor scenes. The KITTI-3D dataset includes different vehicles in outdoor environments and was captured with a synchronized RGB camera and a LIDAR sensor.

Quantitatively, we compare our framework with the most closely related approaches, DSP-SLAM~\cite{wang2021dsp} and Node-SLAM~\cite{sucar2020nodeslam}. DSP-SLAM estimates the object model and 7-DoF pose with deterministic 3D and 2D loss functions. To fairly compare, we extend the DSP-SLAM open source code base to output a 9-DoF object pose, which we refer to as \textit{DSP-SLAM*}. Node-SLAM is not open-sourced, so we implement it from scratch and follow its original design of using only the 2D rendering loss. Node-SLAM measures the rendered depth uncertainties by computing the sampled depth variance along the camera ray, and minimizes the NLL loss. In our experiments, we notice that the NLL loss has limited numerical stability, which makes it difficult to find converging hyperparameters. We therefore implement Node-SLAM with the energy score as a loss, named as \textit{Node-SLAM*}. Since our optimization framework is adaptable to any 2D rendering function, we additionally implement our approach with two variants by exploiting other rendering functions: deterministic rendering (from DSP-SLAM) and sampling-based rendering (from Node-SLAM). We refer to them as \textit{Ours w/ Det2D} and \textit{Ours w/ Samp2D}, respectively. For all the baselines, variants, and our method, we use DeepSDF as the shape model and Adam~\cite{kingma2014adam} as the optimizer. We provide the same inputs, including depth, 2D masks, camera poses, initial object poses, and latent code for all comparisons.

\begin{figure}[t]
\centering
  \includegraphics[width=0.95\linewidth]{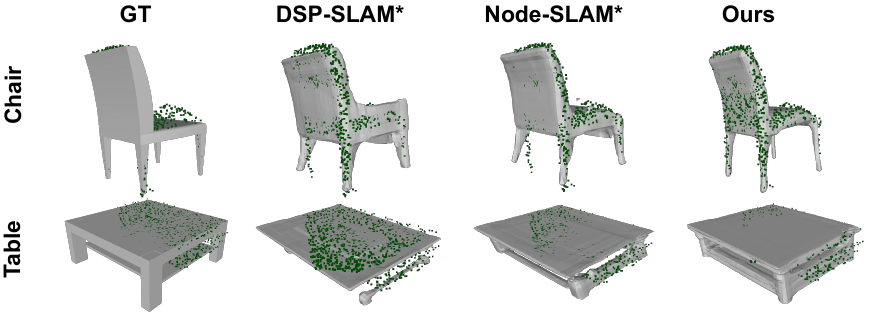}
  \vspace{-0.4\baselineskip}
\caption{Qualitative results on the ScanNet dataset.}
\label{fig_qualitative_scanet}
\vspace{-0.8\baselineskip}
\end{figure}

\begin{table}[t!]
\resizebox{1.0\columnwidth}{!}{
\begin{tabular}{@{}clcccccc@{}}
\toprule
\multirow{2}{*}{\textbf{Views}} & \multirow{2}{*}{\textbf{Methods}} & \multicolumn{3}{c}{\textbf{Chair}}                                                                                   & \multicolumn{3}{c}{\textbf{Table}}                                                                                   \\ \cmidrule(l){3-8} 
                                &                                   & \multicolumn{1}{c}{\textbf{9-DoF Pose}} & \multicolumn{1}{c}{\textbf{IoU>0.25}} & \multicolumn{1}{c}{\textbf{CD<0.2}} & \multicolumn{1}{c}{\textbf{9-DoF Pose}} & \multicolumn{1}{c}{\textbf{IoU>0.25}} & \multicolumn{1}{c}{\textbf{CD<0.2}} \\ \midrule
\multirow{5}{*}{1}              & Node-SLAM*                             & 0.200                                  & 0.487                                 & 0.555                               & 0.098                                  & 0.252                                 & 0.359                               \\
                                & DSP-SLAM*                              & 0.224                                  & 0.603                                 & 0.607                               & 0.105                                  & 0.329                                 & 0.380                               \\
                                & Ours w/ Det2D                          & {\ul 0.225}                            & {\ul 0.631}                           & {\ul 0.636}                         & 0.103                                  & \textbf{0.339}                        & 0.380                               \\
                                & Ours w/ Samp2D                          & \textbf{0.226}                         & 0.621                                 & 0.633                               & \textbf{0.140}                         & 0.316                                 & \textbf{0.439}                      \\
                                & Ours                              & 0.214                                  & \textbf{0.635}                        & \textbf{0.655}                      & {\ul 0.123}                            & {\ul 0.332}                           & {\ul 0.395}                         \\ \midrule
\multirow{5}{*}{2}              & Node-SLAM*                             & 0.202                                  & 0.505                                 & 0.576                               & 0.117                                  & 0.248                                 & 0.374                               \\
                                & DSP-SLAM*                              & 0.228                                  & 0.637                                 & 0.634                               & 0.130                                  & 0.324                                 & 0.382                               \\
                                & Ours w/ Det2D                          & \textbf{0.236}                         & {\ul 0.660}                           & {\ul 0.662}                         & 0.119                                  & {\ul 0.345}                           & 0.396                               \\
                                & Ours w/ Samp2D                          & {\ul 0.231}                            & 0.648                                 & \textbf{0.666}                      & \textbf{0.144}                         & 0.328                                 & \textbf{0.432}                      \\
                                & Ours                              & 0.224                                  & \textbf{0.669}                        & 0.656                               & {\ul 0.138}                            & \textbf{0.360}                        & {\ul 0.430}                         \\ \midrule
\multirow{5}{*}{3}              & Node-SLAM*                             & 0.215                                  & 0.503                                 & 0.601                               & {\ul 0.140}                            & 0.246                                 & 0.390                               \\
                                & DSP-SLAM*                              & 0.239                                  & 0.657                                 & 0.647                               & 0.134                                  & 0.338                                 & 0.398                               \\
                                & Ours w/ Det2D                          & \textbf{0.253}                         & \textbf{0.696}                        & \textbf{0.695}                      & 0.128                                  & \textbf{0.349}                        & 0.417                               \\
                                & Ours w/ Samp2D                          & {\ul 0.252}                            & 0.675                                 & {\ul 0.692}                         & \textbf{0.143}                         & {\ul 0.339}                           & {\ul 0.426}                         \\
                                & Ours                              & 0.226                                  & {\ul 0.682}                           & 0.662                               & 0.132                                  & 0.332                                 & \textbf{0.449}                      \\ \bottomrule
\end{tabular}
}
\vspace{-0.5\baselineskip}
\caption{Quantitative results on the ScanNet dataset.}
\vspace{-1.0\baselineskip}
\label{table_quantitative_scanet}
\end{table}

\subsection{Results on ScanNet Dataset}
We perform the evaluation on two common categories, chair and table, with all video sequences. We use 954 chair instances and 256 table instances from the ScanNet dataset~\cite{dai2017scannet}. For each object, it provides the ground truth CAD model and 9-DoF object pose~\cite{avetisyan2019scan2cad}. We visualize the reconstruction results in Figure~\ref{fig_qualitative_scanet}. Compared to the baselines, our approach reconstructs the objects with far fewer artifacts and better alignment to the point cloud.

For the quantitative evaluation, we calculate the correct detection rate with three metrics: absolute 9-DoF pose, Intersection over Union (IoU) and Chamfer Distance (CD). The 9-DoF pose metric considers an object reconstruction correct if the pose error is within thresholds of 0.2-meter translation, 20-degree rotation, and 20\%-scale. We use IoU with 0.25 and CD with 0.2 meters as thresholds for the other two metrics. For each category, we evaluate the correct detection rate with 1, 2, and 3 viewpoints. We show quantitative results in Table~\ref{table_quantitative_scanet}. One of our approaches ranks first on all numbers of views across all three reconstruction metrics. This shows the benefits of our uncertainty-aware framework for reconstructing the object shape and relative poses. It is noteworthy that our approach exceeds other variants on the single-view test set, where the task is challenging due to incomplete depth and occlusions. However, with more available views, the performance improvement is less noticeable when compared with other variants. This is likely due to using a uni-modal Gaussian distribution to approximate the actual depth distribution, which is usually multi-modal. This phenomenon is more obvious in multi-view cases. To overcome this problem, we consider using a more advanced modelling technique, such as Gaussian Mixture Model~\cite{fu2021multi,yang20236d}, as future work.

\begin{figure}[t]
\centering
  \includegraphics[width=0.98\linewidth]{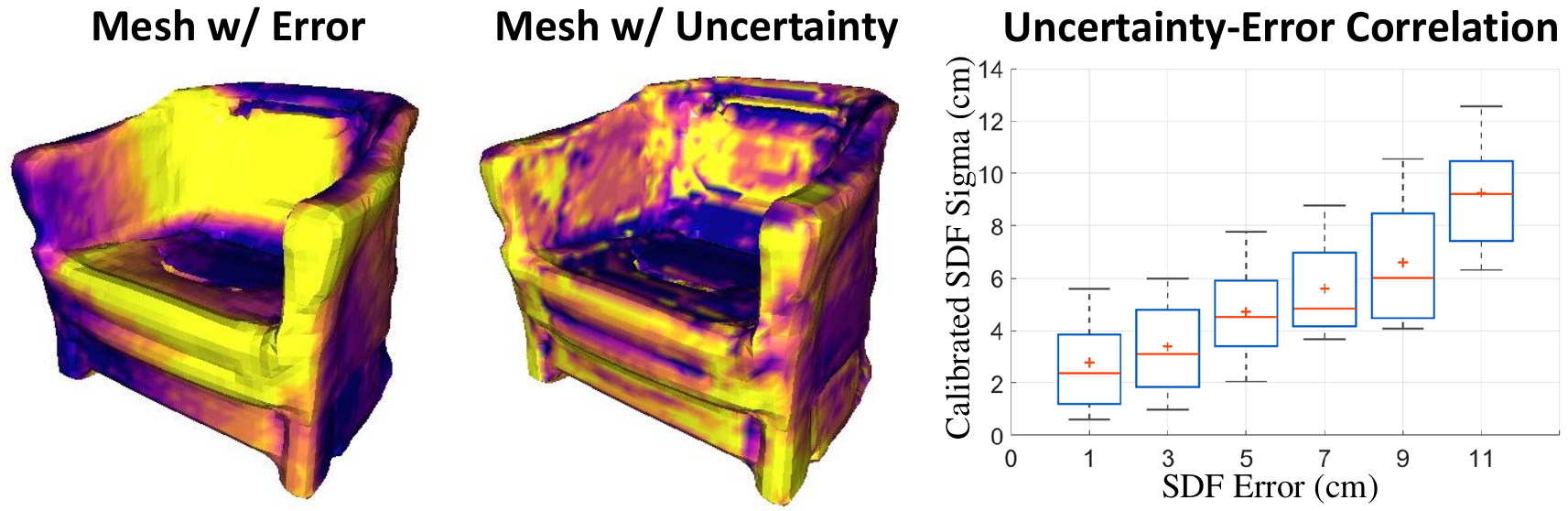}
  \vspace{-0.5\baselineskip}
\caption{Uncertainty and error correlation. Our estimated uncertainty correlates with the true reconstruction errors.}
\vspace{-0.9\baselineskip}
\label{fig_uncertainty}
\end{figure}

\begin{table}[t!]
\centering

\begin{tabular}{@{}cccc@{}}
\toprule
\textbf{Methods} & \multicolumn{1}{l}{\textbf{Ours w/ Det2D}} & \multicolumn{1}{l}{\textbf{Ours w/ Samp2D}} & \multicolumn{1}{l}{\textbf{Ours}} \\ \midrule
Chair            & 0.738                                      & 0.740                                       & \textbf{0.749}                    \\
Table            & 0.661                                      & 0.668                                       & \textbf{0.742}                    \\ \bottomrule
\end{tabular}
\vspace{-0.6\baselineskip}
\caption{Uncertainty evaluation on the ScanNet dataset with the Pearson correlation score metric.}
\vspace{-0.9\baselineskip}
\label{table_uncertainty}
\end{table}

Although the accuracy improvement on multi-view cases is less obvious when using our rendering function, its probabilistically complete rendering process brings us a better ability to estimate the uncertainty in pose and shape. To evaluate the estimated uncertainties, we first generate GT depth data by sampling $10K$ points from the GT object model under the world frame, $W$. For each point, we transform it to the object frame, $O$, using the estimated object pose, $\boldsymbol{\xi}_{ow}$, and compute the SDF mean and variance with the estimated latent code, $\boldsymbol{z}$. A well-estimated variance should correlate to the actual SDF error. Figure~\ref{fig_uncertainty} first shows a qualitative example of our estimated SDF variance, which correlate well with the true SDF errors. Quantitatively, we evaluate this correlation against our variants with the Pearson correlation score. We evaluate on well-reconstructed objects under the 3-view setup, including 145 chairs and 17 tables. Table~\ref{table_uncertainty} shows the uncertainty evaluation on the ScanNet dataset. Our approach achieves higher Pearson scores than the other two variants, demonstrating the effectiveness of our probabilistic rendering for uncertainty estimation.

\subsection{Results on KITTI Dataset}
We use the KITTI-3D object detection dataset~\cite{geiger2012we} to evaluate the system performance in the outdoor environment. We obtain initial object poses using the PointPillars 3D detector~\cite{lang2019pointpillars} and acquire object masks from the Mask2Former segmentation algorithm~\cite{cheng2021mask2former}. To investigate the upper bound of reconstruction accuracy of each method, we consider an object for evaluation only if its initial 3D detection error is within thresholds of 1.0-meter translation and 20-degree heading. In total, we performed the evaluation on 253 vehicles from the validation set.

\begin{figure}[t]
\centering
  \includegraphics[width=0.97\linewidth]{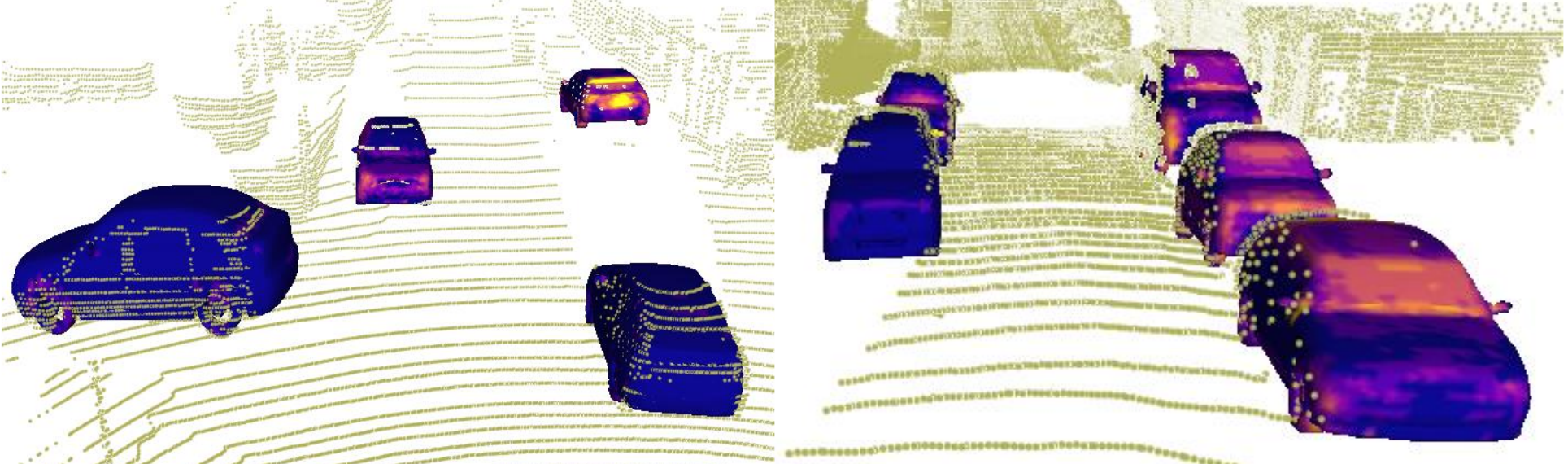}
  \vspace{-0.4\baselineskip}
\caption{Qualitative results on KITTI-3D.}
\vspace{-0.8\baselineskip}
\label{fig_qualitative_kitti}
\end{figure}

\begin{table}[t]
\centering
\begin{tabular}{@{}lccc@{}}
\toprule
\textbf{Methods} & \textbf{Mean IoU} & \textbf{IoU\textgreater{}0.6} & \textbf{IoU\textgreater{}0.75} \\ \midrule
Node-SLAM*       & 0.677             & 78.6                          & 21.4                           \\
DSP-SLAM*        & 0.690             & 80.2                          & 25.7                           \\
Ours w/ Det2D       & 0.721             & 88.9                          & 42.3                           \\
Ours w/ Samp2D      & \textbf{0.741}    & \textbf{93.9}                 & {\ul 53.7}                     \\
Ours      & {\ul 0.738}       & {\ul 90.5}                    & \textbf{54.9}                  \\ \bottomrule
\end{tabular}
\vspace{-0.5\baselineskip}
\caption{Quantitative results on KITTI-3D.}
\vspace{-0.8\baselineskip}
\label{fig_quantitative_kitti}
\end{table}

Since object shape annotation is not available, we evaluate the performance with the IoU metric in  Table~\ref{fig_quantitative_kitti}. Our approaches, including two variants, exceed the baselines, \textit{DSP-SLAM*} and \textit{Node-SLAM*}, by a large margin. We also notice that our approach exceeds other variants when using a more strict metric (IoU>0.75). Further qualitative results in Figure~\ref{fig_qualitative_kitti} demonstrate the high performance when recovering the shape and relative poses for different vehicles, even with partial Lidar observations.

\subsection{Computation Analysis}
All experiments were performed on a 16GB V100 GPU. For our approach, it costs 0.2s with each iteration. Depending on the task, the shape resolution (currently $64^3$), the sampling number, and the iteration steps can be modified to trade-off between the efficiency and accuracy.




\section{CONCLUSION}
In this work, we presented an object-level mapping approach that can recover the 3D dense models and relative poses for unknown objects. The core idea of our approach is leveraging a learnt category-specific shape prior to formulate an uncertainty-aware optimization framework. We introduce two probabilistic loss functions that model the uncertainties of shape and pose in the optimization. We compare our approach against the state-of-the-art approaches on challenging real-world datasets, ScanNet and KITTI-3D. The results demonstrate that our approach can reconstruct higher-quality object-level maps. Moreover, our estimated uncertainties accurately correlate with the true errors of our estimated object shapes and poses, which is valuable for many downstream robotic applications. This work represents an important step toward our future development of an uncertainty-aware object-level SLAM system that jointly estimates camera poses and actively selects camera viewpoints for building detailed object-level maps in the open world.






\bibliographystyle{ieeetr}
\bibliography{bibliography.bib}

\end{document}